\documentclass[final]{cvpr}

\usepackage{epsfig}
\usepackage{graphicx}
\usepackage{amsmath}
\usepackage{amssymb, eucal, nicefrac}
\usepackage{multirow}
\usepackage{threeparttable}
\usepackage{booktabs}
\usepackage{dcolumn}
\usepackage{float} 
\usepackage{subfig}
\usepackage{xfrac}
\usepackage[margin=6pt,font=small,labelfont=bf,labelsep=endash,tableposition=top]{caption}

\usepackage{xcolor}		
\usepackage[colorlinks = true,
            linkcolor = purple,
            urlcolor  = blue,
            citecolor = cyan,
            anchorcolor = black]{hyperref}

\def\cvprPaperID{2643} %
\def\confYear{CVPR 2021}
\begin{document}

\title{Generic Perceptual  Loss for Modeling Structured Output Dependencies
}

\author{Yifan Liu$ ^1$,
~
~
~
           Hao Chen$ ^1$,    
~
~
~
Yu Chen$ ^2$,
~
~
~
   Wei Yin$ ^1$,
~
~
~
        Chunhua Shen$ ^{1,3}$\thanks{C. Shen is the 
corresponding author.}
        \\
        [0.2cm]
        $ ^1$ The University of Adelaide%
        ~~~~~~~
        $ ^2$ Automind %
        ~~~~~~~
        $ ^3$ Monash University, Australia
}

\maketitle
\begin{abstract}
The perceptual loss has been widely used as an 
effective loss term in %
image synthesis 
tasks including 
image super-resolution \cite{ledig2017photo}, %
and style transfer \cite{johnson2016perceptual}.
It was believed that the success  
lies in the  high-level perceptual feature representations extracted
from CNNs \textit{pretrained} with a large set of 
images.
Here we reveal that, what matters is the network structure instead of the trained weights. 
Without any learning, 
the structure of a deep network is sufficient to capture the dependencies  
between multiple levels of variable statistics using multiple layers of CNNs.
This insight removes the requirements of pre-training and 
a particular network structure (commonly, VGG) that are previously assumed for the perceptual loss, thus 
enabling a significantly wider range of applications.  
To this end, we  demonstrate that a randomly-weighted deep CNN can
be used to model the structured dependencies of outputs. 
On a few dense per-pixel prediction tasks such as semantic segmentation, 
depth estimation and instance segmentation, we show improved results of using 
the extended randomized perceptual loss, compared to the baselines using pixel-wise loss alone. 
We hope that this simple, extended perceptual loss may serve as a generic 
structured-output 
loss that is applicable to most structured output learning tasks.

\end{abstract}

\section{Introduction}

Dense pixel-wise prediction tasks  represent 
the most important category of computer vision problems, ranging from low-level image processing such as denoising,  super-resolution, through mid-level tasks such as stereo matching, to high-level understanding such as semantic$/$instance segmentation.
These tasks are naturally 
structured output learning problems since the prediction variables  
often depend on each other. 
The pixel-wise loss serves as the unary term for these tasks.
Besides,
the perceptual loss~\cite{johnson2016perceptual} was introduced
to capture perceptual information by measuring 
discrepancy
in high-level 
convolutional 
features extracted from CNNs.
It has been successfully used in various low-level image processing tasks, such as style transfer, and super-resolution \cite{johnson2016perceptual}.

Previous works %
assume 
that the perceptual loss benefits from the high-level perceptual features extracted from CNNs pretrained with a large set of %
images
(\textit{e.g.}, VGG \cite{simonyan2014very} pretrained on %
the
ImageNet %
dataset.
Relying 
on this assumption, the perceptual loss is limited to \textit{a %
specific 
network structure} (commonly, VGG) with \textit{pre-trained} weights, which 
is not able to take arbitrary signals as the input. 
In this work, we reveal that, contrary to this belief, the success of the perceptual loss is not necessarily dependent on the ability of a \textit{pretrained} CNN in extracting high-level perceptual features. 
Instead, \textit{without any learning}, the structure of a multi-layered CNN is sufficient to capture a large amount of interaction statistics for various output forms. 
We argue that what matters is the deep network architecture rather than 
the pretrained weights.

\begin{figure*}[htb]
\centering  
\resizebox{0.99\linewidth}{!}{\subfloat[Ground Truth]{
\includegraphics[width=0.2\textwidth]{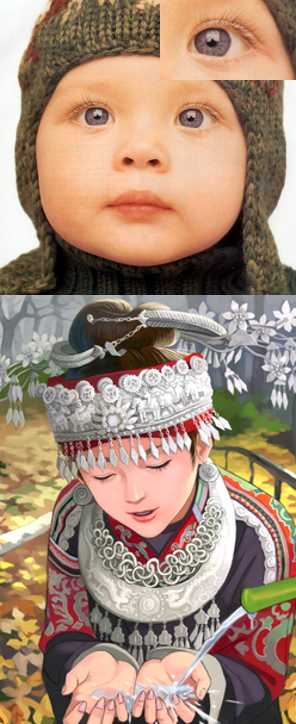}
\label{subfig:3a}}
\quad
\subfloat[Bicubic]{
\includegraphics[width=0.2\textwidth]{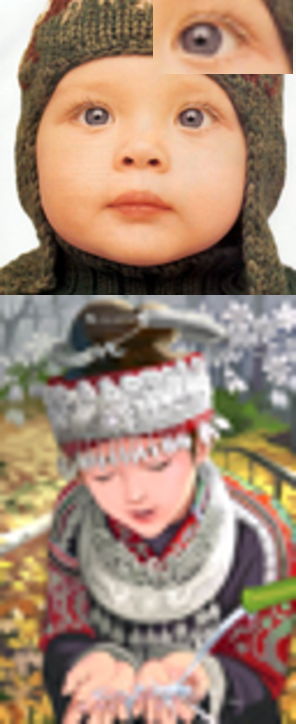}
\label{subfig:3b}}
\quad
\subfloat[Pixel-wise Loss alone~\cite{ledig2017photo}]{
\includegraphics[width=0.2\textwidth]{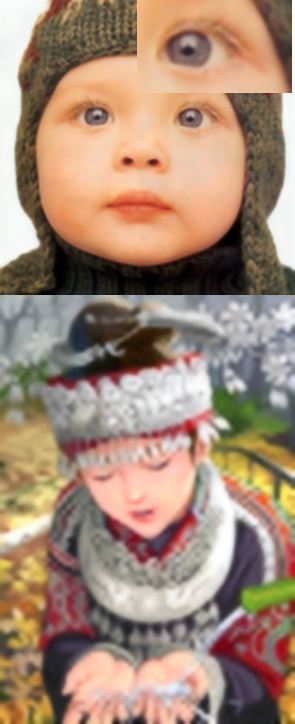}
\label{subfig:3c}}
\quad
\subfloat[w.\ Pretrained VGG~\cite{ledig2017photo}]{
\includegraphics[width=0.2\textwidth]{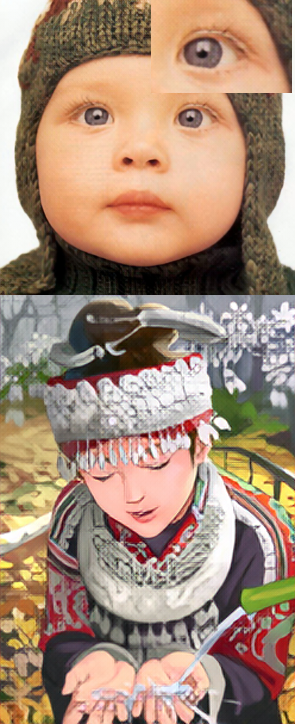}}
\label{subfig:3d}
\quad
\subfloat[w.\ Random VGG]{
\includegraphics[width=0.2\textwidth]{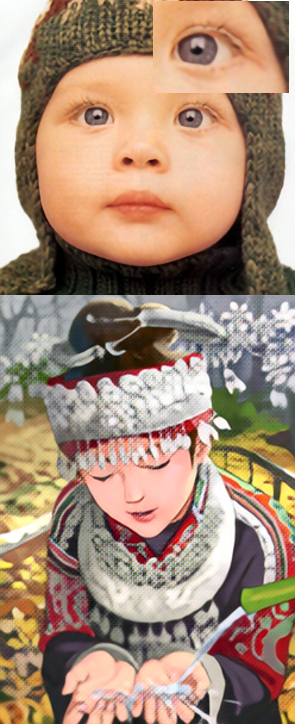}
\label{subfig:3e}}
}
\caption{%
Super-resolution results of the pilot experiments (super-resoled from 4$\times$ down-scaled images). (a) Ground truth high-resolution images. (b) Bi-cubic up-sampling. (c) SRResNet~\cite{ledig2017photo} trained with the per-pixel loss. (d) SRResNe trained with the per-pixel loss and perceptual loss with a pretrained VGGNet. (e) SRResNet trained with the per-pixel loss and perceptual loss with a randomly-weighted VGGNet.
We can see that %
 the 
perceptual loss %
improves image quality. Besides, formulating the perceptual loss with a pre-trained network and a randomly-weighted network produces
\textit{on par}
results.
}
\label{fig:sr}
\vspace{-1em}
\end{figure*}

To verify the statement, we conduct a pilot experiment on image super-resolution. Apart from using the pretrained VGG net for perceptual loss, we use a randomly-weighted 
network. 
The results with the randomly-weighted network are 
\textit{on par} with that of 
the pretrained %
VGG, 
which are both visually 
improved 
than %
using the per-pixel loss alone (see Figure~\ref{fig:sr}).
This 
indicates 
that
the pretrained weights---%
previously assumed for the perceptual loss---is not 
essential to the success of the perceptual loss.
We may conclude from this experiment  that 
it is  
the deep network structure,
rather than learnt 
weights, plays %
the core role.

Given a target $y$ or a 
prediction
$\hat y$ as an input, a randomly-weighted 
network $ f( \cdot ) $ can work as a function to explore hierarchical dependencies between variable statistics through the convolution operations in multiple layers. 
Thus, 
a generic perceptual loss for structured output learning can be computed by comparing
the 
discrepancy 
between $ f^j(y)$ and $ f^j (\hat y)$. Here $ j $ indexes a particular layer of the 
network $ f(\cdot )$.
Thus, 
this enables the perceptual loss\footnote{Here we still use the notion of `perceptual' as it was firstly introduced in \cite{johnson2016perceptual} even though broadly this loss is more about capturing inter-dependencies in variables, instead of extracting perceptual features.
}
to be applied to a wider range of structured output learning tasks. 

Structured information
is important in dense per-pixel prediction problems, such as semantic segmentation
\cite{lin2016efficient}, depth estimation and instance segmentation
\cite{liu2019structured}. 
For example, the pairwise term in Markov Random Filed
is 
complementary to the unary term, which
defines pairwise compatibility and in general improves prediction accuracy  
especially when the unary term alone is %
not sufficient. 
The %
proposed 
generic perceptual loss can be easily applied to these dense prediction tasks, with \textit{no computation overhead} during inference. 
Also, as now pre-training with labelled data is not required, it is 
straightforward 
to explore the effectiveness of using 
various
network structures---not necessary the VGG---to model the  dependency between output variables.

Experimental results on various structured output learning tasks with different network structures show that the generic perceptual loss %
benefits the training, and consistently 
achieves %
improved 
performance compared to the baselines using pixel-wise loss term alone. 
We also 
provide
detailed comparisons and analysis on the impact of  initialization schemes and architectures of the perceptual loss network.

In summary, our main contributions %
are as follows.
\begin{itemize}
\itemsep -0.1cm
    \item 
    
 We reveal that the success of the perceptual loss is not dependent on the pretrained CNN weights. 
 Without any %
 learning, 
 the structure of a deep network is sufficient to capture the dependencies between 
 multiple 
 levels of variable statistics using multiple layers of CNNs. 

    \item 
    
 We apply the generic perceptual loss %
 to a few structured output learning tasks, including semantic segmentation, depth estimation and instance segmentation. 
We consistently improve the performance of baseline models. 
    \item 

 We investigate how the initialization and the network structures 
 may
 affect the performance of the %
 proposed
 perceptual loss. A reliable initialization approach is %
 designed based on the analysis.

    \item 
    
 This proposed simple 
 perceptual loss may %
 serve as a generic structured-output loss that is applicable to most structured output learning tasks in computer vision.

\end{itemize}

\section{Related Work}
\noindent{\bf Perceptual loss.} %
Early 
works~\cite{ledig2017photo,johnson2016perceptual%
} generate high-quality images using perceptual loss functions, which consider the %
discrepancy 
between deep features, not only the pixels. 
Gatys \etal~\cite{Gatys_2016_CVPR} found that a pretrained VGG architecture can be used to as a loss function for %
style transfer. 
They attribute the success to the ability of the trained filters in learning certain features which are coincident to human perception. 
Johnson \etal~\cite{johnson2016perceptual} further formulate  the perceptual loss as an extra %
loss term 
for the deep neural network. %
There, the context/perceptual loss  is the Euclidean distance between feature representations. 
Inclusion of 
the perceptual loss achieves visually improved results on style transfer and super-resolution.
While doing other image synthesis tasks, such as super-resolution%
,
colorizing,
and other image generation tasks \cite{isola2017image%
},  the perpetual loss often refers to this context loss as shown
 in Eq.~\eqref{context},
\begin{equation}
\label{context}
  \ell_{feat}^{\phi,j}(\hat y, y) = 
  \frac1{C_jH_jW_j}\|\phi_j(\hat y) - \phi_j(y)\|_2^2,
\end{equation}
where $y$ and $\hat y$ are the targeted images and synthesis images. $\phi_j$ represents the perceptual function 
which outputs the activation
of the $j$th layer in the perceptual loss network. $ C_j $,
$ H_j $, $ W_j $ are the dimensions of the tensor feature map. 

Recently, %
a perceptual loss
was also introduced to depth estimation task~\cite{wang2020cliffnet}. 
The authors argue that 
the embedding spaces should be designed for particular relevant tasks, \textit{i.e.}, depth-based scene classification and depth reconstruction. 
Thus, they have pretrained the perceptual loss network on RGBD datasets and 
failed to realize that pretraining is not compulsory, as we show here. 
\noindent{\bf Representations with random weights.}
A few methods have discussed the untrained, randomly-weighted CNNs. 
Researchers found that networks with random weights can extract useful features as the classification accuracy with these features is higher than 
random guesses. 
He \etal~\cite{he2016powerful} use generative models with the constraints from untrained, randomly-weighted network for deep visualization tasks. They found that during optimization for a style transfer task, the perceptual loss with a proper weight scale can work well with untrained, randomly-weighted networks, and generate competitive results as prior work~\cite{Gatys_2016_CVPR} with pre-trained weights. 
Mongia \etal~\cite{mongia2017random} prove why one-layer CNNs with random weights can successfully generate textures. 
The randomly-weighted networks are also employed in 
unsupervised learning 
 \cite{wang2019unsupervised} 
and 
reinforcement learning %
\cite{gaier2019weight}. 
Another relevant work is deep image prior \cite{DIP}  where 
a randomly-initialized neural network is used as a ``handcrafted prior'' with excellent results in image reconstruction tasks such as denoising, super-resolution, and inpainting. 
These works provide a way to study network architectures without any learning, and also exploit the randomness as a useful feature.
Here, we explore the ability of the randomly-weighted network in investigating hierarchical dependencies between variable statistics. The perceptual loss with a randomly-weighted network works as a %
useful loss term 
on various structured output learning tasks.

\noindent{\bf Dense prediction.}
Dense prediction is a family of fundamental problems in computer vision, which learns a per-pixel mapping from input images to output structures, including semantic segmentation~\cite{%
yu2020representative}, depth estimation~\cite{wei2019enforcing%
}, object detection~\cite{tian2019fcos}, \textit{etc.}
As extensively studied in the literature,  taking the inter-dependency between output variables into account during training and$/$or inference often improves the accuracy. Thus    
structured information is important for these tasks. 

In this work, we %
demonstrate 
that a randomly-weighted network can implicitly capture the structural information with its natural architecture and internal convolution operations. 
The performance of these dense prediction tasks can be enhanced by simply enforcing a perceptual loss from a randomly-weighed network. This is achieved \textit{without any learning} on the perceptual loss network, and \textit{no further computation cost} for inference is required.

\section{Our Method}
The observations from the pilot experiments in Figure~\ref{fig:sr} suggest that the network structure, instead of the pretrained weights, contributed to the success of the perceptual loss. More details can be found in the supplementary materials. In this section, we first extend the perceptual loss with randomly-weighted networks to some structured output learning tasks. Then, we analyze the devils in the weight initialization and design an appropriate way for an effective initialization.

\subsection{Perception Loss for Structured Output Prediction}
If the randomly-weighted network has the ability in capturing structured information, it should also be able to help dense prediction problems.
In addition, as the pretrained weights learnt with a large number of samples are not required, it is easier to apply this regularization on any task and to compare the performance with different perceptual loss networks.
We start with the commonly used VGGNet-like structure as the perceptual loss network
. We denote the number of convolutional layers between max pooling downsample operations with $N_{1}, N_{2}, \dots, N_{k}$, where $k$ is the number of blocks.

\noindent\textbf{Semantic segmentation.}
Semantic segmentation is a typical dense prediction problem, where a semantic label is assigned to each pixel in an input image. 
A segmentation network as $\mathcal{S}$ takes an input image $\mathbf{I} \in \mathbb{R}^{W \times H \times 3}$ and predicts a segmentation map $\hat y=\mathcal{S}(\mathbf{I})\in \mathbb{R}^{W \times H \times C}$. The output channel of the segmentation network $C$ equals to the number of the pre-define object classes.
Conventional methods usually employ a per-pixel cross-entropy loss.
The correlations among pixels are neglected in the cross-entropy loss. Therefore, the perceptual loss can work as a complementary to the per-pixel loss for capturing the structured information.

To extend the perceptual loss to semantic segmentation, we use the estimated segmentation map or the ground-truth one as the input to the perceptual loss network, and get the embedded structured feature after several CNN layers. 
The mean square error is used to minimize the distance between the structured features of prediction and the learning target. 
The softmax output $\hat y$ has a domain gap with the one-hot ground truth $y$, which make the perceptual loss hard to converge.
To solve this problem, we follow recently knowledge distillation methods~\cite{%
liu2019structured} to generate soft labels $y_t$ by a large teacher net as the learning targets. 
The total loss is then defined as:
\begin{equation}
\def\bn{ { \boldsymbol n  } }
\label{eq:vnl}
    \ell_{\rm seg} = \ell_{ce}(\hat y,y)+\lambda \cdot  \ell_{r}^{\phi_r}(\hat y, y_t),
\end{equation}
where $\phi_r$ represents the perceptual loss network initialized with random weights and we set $\lambda$ as $0.1$ in all experiments.
\noindent\textbf{Depth estimation.}
Monocular depth prediction~\cite{wei2019enforcing, fu2018deep} is a regression problem, which predicts the per-pixel real-world distance from the camera imaging plane to the object captured by each pixel in a still image $\mathbf{I}$. 
We use VNL proposed by Yin~\etal~\cite{wei2019enforcing} as a baseline model. The pixel-level depth prediction loss and the virtual normal loss 
are used to supervise the network. 
The network outputs a predicted depth map $\hat{d}\in \mathbb{R}^{W \times H \times 1}$. 
Therefore, the input channel of the perceptual loss net equals to $1$. 
As the ground-truth depth map follows the same statistical distribution as the estimations, the target and the estimation can be directly used as inputs to the perceptual loss network.
The network with virtual normal loss, as a strong baseline, minimizes the difference of a manually defined geometry information between the prediction and the ground truth, \textit{i.e.},  the direction of the normal recovered with three samples. 
The perceptual loss can still capture extra structured information when combined with $\ell_{vn}$.

\noindent\textbf{Instance segmentation.}
Instance segmentation is one of the most challenging computer vision
tasks, as it requires the precise per-pixel object detection and semantic segmentation simultaneously. 
Recently, one stage methods achieve promising performance~\cite{tian2020conditional,chen2020blendmask,wang2020solov2}, making the pipeline more elegant and easier to implement. 
The mask and classification logits are predicted for each pixel in the feature space. 

We follow CondInst~\cite{tian2020conditional}, a state-of-the-art one-stage method, as a strong baseline. 
In the training procedure, 
CondInst consists of a bounding box detection head and a mask head.
The detection head also includes a controller, which is dynamically applied to the mask head for different instances.
In this way, it will produce single channel segmentation masks for each instance in the training batch with the shape of $\sfrac{1}{4}W \times \sfrac{1}{4}H$. If there are $n$ predicted instances in the training batch, the input of the perceptual loss network is then $n \times 1 \times \sfrac{1}{4}W \times \sfrac{1}{4}H$.
Similar to semantic segmentation, the soft targets are generated by a teacher network. 

\subsection{Devils in the Initialization}
\label{sec:init}
The initialization of the randomly-weighted network affects the performance of the generic perceptual loss. 
As we employ the structured predictions as the input to the perceptual loss network and produce an embedding,
an inappropriate initialization %
may
lead to an unstable %
results. 
It is important to guarantee that each layer is bounded by a Lipschitz constant close to $1$, so that the gradient generated by the random network will not explode or vanish.
Here we investigate the gradient scales in the random network and derive a robust initialization method by following~\cite{he2015delving}.
For a dense prediction task, we have the prediction results (\textit{e.g.}, segmentation map) $\mathbf{Y'}$, and also its ground-truth or soft targets $\mathbf{Y}$. 
Our random-weight perceptual loss network transforms $\mathbf{Y'}$ and ${\bf Y}$ into two embeddings $\mathbf{E'}$ and $\mathbf{E}$.
The generic perceptual loss is
computed as 
the %
discrepancy 
between $\mathbf{E'}$ and $\mathbf{E}$, 
\begin{equation}
\ell_{r}=\|\mathbf{E'}-\mathbf{E}\|_2^2.
\end{equation}
Suppose that $\mathbf{e}_l$ is the response activation values related to the corresponding $k \times k$ pixels in the convolution operation with kernel size $k$. We %
have:
\begin{equation}\label{eq:forward}
\mathbf{e}_l=\mathbf{W}_l\mathbf{y}_l + \mathbf{b}_l.
\end{equation}
Here, $l$ is the index of a layer. $\mathbf{y}_l$ is the input vector with $n_l=k^2c_l$ elements, where $c_l$ is the input channel of this layer.  $\mathbf {W}_l $ is a $d_l$-by-$n_l$ matrix, where $d$ is the number of random initialized filters in this layer. With a deep convolutional network, we have
$\mathbf{y}_{l}=f(\mathbf{e}_{l-1})$, where $f(\cdot ) $ is the ReLU activation function. We also have $c_l = d_{l-l}$. 
If we initialize %
$w_{l}$ with a symmetric distribution around zero and $b_{l}=0$, then $e_{l}$ has zero mean and has a symmetric distribution around zero. Following~\cite{he2015delving}, we %
compute 
the variance of the output embedding after $ L $ 
layers:
\begin{equation}\label{eq:prod_fw}
\emph{Var}[e_{L}]=\emph{Var}[e_{1}]\left(\prod_{l=2}^{L}\frac{1}{2}n_l \emph{Var}[w_{l}]\right).
\end{equation}
The variance of the discrepancy is upper bounded by the same factor:

\begin{equation}
    \begin{aligned}
&\emph{Var}[(e'_{L}-e_{L})]\\
&=\emph{Var}[e'_{L}]+\emph{Var}[e_{L}]-2\emph{Cov}[e'_{L},e_{L}] \\
&\le\emph{Var}[e_{l}]+\emph{Var}[e'_{l}])(\prod_{l=2}^{L}\frac{1}{2}n_l \emph{Var}[w_{l}].
    \end{aligned}
\end{equation}
When $L$ becomes extremely large, the product $\prod_{l=2}^{L}\frac{1}{2}n_l \emph{Var}[w_{l}]$ %
vanishes or explodes if $\frac{1}{2}n_l \emph{Var}[w_{l}] \neq 1$. Therefore we initialize each layer using a zero-mean Gaussian distribution with a standard deviation (std) of $\sqrt{2/{n_l}}$ %
as in \cite{he2015delving}. Before the optimization of the task network, the correlation between $  Y$ and ${Y}'$ is very small. Therefore, the scale of the $\theta$ is small. In the training process, the weights of the random network ($\mathbf{W}$) are fixed, but %
$Y'$ %
becomes closer to $Y$ as the task network is optimized. 
Then the covariance becomes close to the variance of the two embeddings and this bound tends to zero.

\section{Experiments}
In this section, we first investigate some interesting questions about the perceptual loss network, and then employ an efficient and effective structure as the perceptual loss network to show its ability in boosting performance in %
a few 
dense prediction tasks, including semantic segmentation, depth estimation and instance segmentation.

\subsection{Discussions}
We share some observations in exploring the capacity of the random weight perceptual loss networks. 
We 
ask a few 
questions including: \textit{Will the trained filters help the perceptual loss in dense prediciton problems? How does the depth/receptive field/multi-scale losses affect the performance? How does the initialization affect the performance?}
Discussions are base on semantic segmentation task with Cityscapes~\cite{Cordts2016Cityscapes} as the training set.
PSPNet~\cite{zhao2017pyramid} with Resent$18$~\cite{He2016DeepRL} as the backbone is used as a baseline model, which is trained with the per-pixel cross-entropy loss.
The soft targets are generated by the PSPNet with Resent$101$~\cite{He2016DeepRL} as the backbone. 
The training settings follow the details in Section~\ref{sec:seg}.
The performance is evaluated on the validation set of Cityscapes with the mean of Intersection over Union (mIoU) as the metric. 

\begin{table}[]
\footnotesize
\centering
\begin{tabular}{ r |c|c}
\hline
P Net &R: mIoU (\%)& T: mIoU (\%)         \\
\hline
             \multicolumn{3}{c}{Non-VGG families} \\
\hline
GoogleNet \cite{szegedy2015going}
             &$68.91$&$68.90$ \\
AlexNet~\cite{krizhevsky2017imagenet}&$69.80$&$69.87$ \\
MobileNetV2~\cite{Sandler2018MobileNetV2IR}&$69.98$&$70.01$ \\
ResNet18~\cite{He2016DeepRL}&$70.16$& $70.14$\\
\hline
             \multicolumn{3}{c}{VGG families} \\
\hline
VGG16~\cite{simonyan2014very}        &  $70.68$ &$70.71$ \\
VGG19~\cite{simonyan2014very}          &$\bf{71.25}$  &$71.19$\\
\hline
\end{tabular}
\vspace{1em}
\caption{Results of the perceptual loss %
for semantic segmentation 
with %
a few different
networks.
`R' indicates that we randomly weight the loss
network. `T' means that we assign the network with the pretrained kernels from ImageNet classification. The baseline model achieves  $69.60\%$ of mIoU.}
\vspace{-1.5em}
\label{tab:structures}
\end{table}

\subsubsection{Training Weights vs. Architecture}
\label{sec:random}
The pretrained filters were considered the key to the success of the perceptual loss. 
Apart from the visualization results on image super-resolution in the previous pilot experiment, we %
show quantitative analysis on structured output learning tasks in this section. 

Taking the semantic segmentation task as an example, we use the same semantic segmentation network and training settings in our experiments, and only change the perceptual loss networks. 
We assign the weights of the perceptual loss network with pretrained kernels to see if the trained filters can help improve the performance. 
Also, we choose different network structures as the perceptual loss network, including VGG families~\cite{simonyan2014very}, ResNet18~\cite{He2016DeepRL}, GoogleNet \cite{szegedy2015going},  AlexNet~\cite{krizhevsky2017imagenet} and MobileNetV2~\cite{Sandler2018MobileNetV2IR}.
The results are shown in Table~\ref{tab:structures}. 
`R' means that the weighs of the perceptual loss network are randomly initialized following Section~\ref{sec:init}. 
`T' means that  we employ the per-trained weights on the ImageNet to initialize the perceptual loss network, and the dense prediction output is transferred from $C$ channels to $3$ using a $1 \times 1$ convolutional layer to fit the pre-trained network structure. 

From the table, we can see that training with random weights and the pre-trained weights %
show almost no 
difference on improvements (difference around 0.02\% to 0.06\% ), but different network structures lead to a larger performance gap (varies from 68.90\% to 71.25\%).

\textit{This indicates that , in this structured output learning task, the trained filters for ImageNet is not the key %
to 
the success of perceptual loss. Meanwhile, the network architecture of the perceptual loss network affects the ability of capturing the structured information.}

Interestingly, we %
observe that 
the VGG families perform %
better than other structures. 
VGG families have been shown unique in %
a few
previous works. 
Researchers find that in style transfer, the VGG structure can work better than ResNet~\cite{ilyas2019adversarial,nakano2019discussion}, and explain %
the reason as the VGGNet is more robust than the ResNet. 
Su \etal~\cite{su2018robustness} %
observe 
that VGG families exhibit high adversarial transfer-ability than other structures. 
However, the theoretical explanation of why VGG structures show a better performance is still not fully investigated in literature.

\begin{table}[t]
\footnotesize 
\centering
\begin{tabular}{ r |l |c|c}
\hline
Percep.\ network & Structure& R: mIoU (\%)&T: mIoU (\%)         \\
\hline
N/A         &1, 1, 1, 1, 1 & $70.88 \pm 0.03$   &N/A\\
VGG11      &1, 1, 2, 2, 2      & $70.18 \pm 0.11$ &$70.21 \pm 0.10$ \\
VGG13    &2, 2, 2, 2, 2       &$70.64 \pm 0.14$   &$70.62 \pm 0.12$ \\
VGG16         &2, 2, 3, 3, 3 &  $70.68 \pm 0.03$ &$70.71 \pm 0.02$ \\
VGG19         &2, 2, 4, 4, 4  &$71.25 \pm 0.04$  &$71.19 \pm 0.07$\\
N/A&3, 3, 4, 4, 4&$70.89 \pm 0.23$&N/A \\
\hline
\end{tabular}
\vspace{1em}
\caption%
{The perceptual loss with variants of VGG as the perceptual loss network (kernel size: $3$). We %
vary
the number of convolutional layers in each block. `R' means random initialization. `T' means initialization with pre-trained weights}
\vspace{-1.5em}
\label{tab:layers}
\end{table}

An independent recent work~\cite{wang2020cliffnet} for depth estimation also shows interesting results by employing a convolutional network to map the structured output into an embedding space.
They argue 
that 
if the perceptual loss network are trained with some highly related tasks, the performance %
is improved. 
So they require additional annotations and try to design an efficient network structure as the perceptual loss network.
Different from their work, we focus on a more general discussion on the perceptual loss, and %
verify that the network architecture plays a more important role than pretrained weights. 
Although training the perceptual loss network on highly relative tasks with additional information may further improve the performance, it is %
not generic to be applied 
to %
various 
tasks and requires extra annotation effort.

\subsubsection{Design of the Perceptual Loss Network}
In the previous section, we have %
shown 
that the network architecture plays a more important role in the perceptual loss. 
As our method does not require pre-training, it is convenient to investigate the performance with various network structures.
Here 
we explore different designs for the randomized perceptual loss network. 
All the experiments are conducted for three times with random initialization, and we report the mean and derivation for each setting.

\noindent\textbf{Impact of depth.} Conventional VGG families have variants with different numbers of layers, such as the most popular VGG16 and VGG19. 
These VGG families %
contains
five convolutional blocks, and a max pooling layer at the end of each block. For each block, the feature dimension is ${64,128,256,512,512}$. 
We change the number of convolutions in each block, as shown in the `Structure' in Table~\ref{tab:layers}. 
For the typical structures in original VGG families, we also report the results with pretrained weight on ImageNet as in Section~\ref{sec:random}. 
All the convolutions' %
kernerl size is  $3$ as in VGGNet.
From Table~\ref{tab:layers}, we can find a consistent conclusion that pretrained kernels %
do not 
help to improve the performance, while the network structures %
exhibit 
a larger impact. 
With the kernel size %
of 
$3$, VGG19 is the most effective perceptual loss network. 
Besides, the structure of `$1,1,1,1,1$' (only 1 conv.\ layer for each block) also %
show good 
performance.
As the training memory and the training time may increase as the perceptual loss network becomes deeper, the structure of `$1,1,1,1,1$' among these settings is the best choice considering both effectiveness and efficiency.

\begin{figure}
    \centering
    \includegraphics[width=0.4595\textwidth]{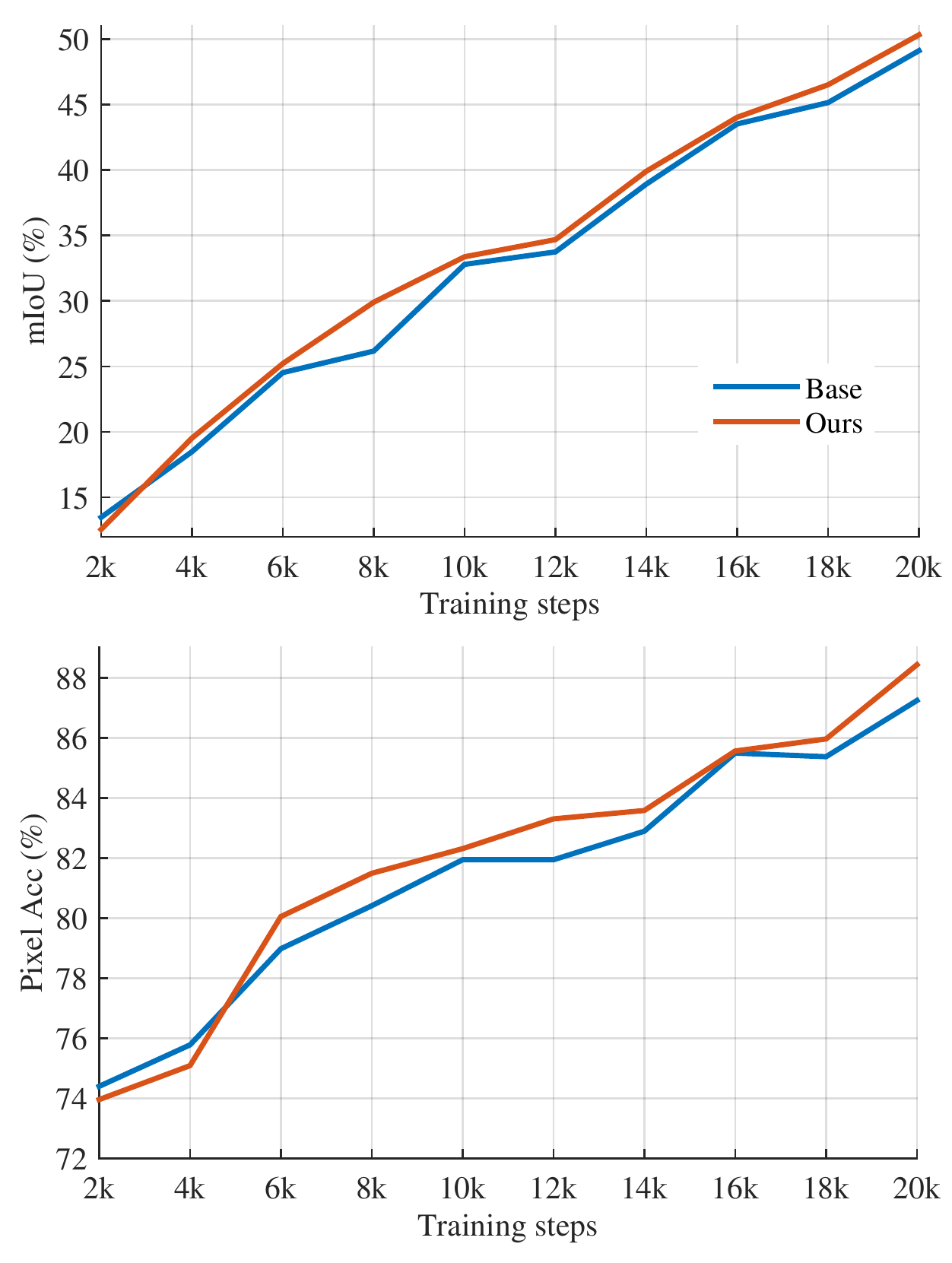}
    \caption{
    Pixel accuracy and %
    mIoU on the validation set during training on the Pascal VOC dataset. `Base' means a semantic segmentation network of PSPRes18 without a perceptual loss. 
    `Ours'
    represents the baseline network trained with a randomized perceptual loss. 
    }
    \label{fig:loss}
\vspace{-0.5em}
\end{figure}

\noindent\textbf{Impact of the receptive field.} 
The randomly-weighted network can capture the structured correlation among local features, as the convolution operation has the ability to %
exploit 
the information %
at multiple 
scales of receptive fields. 
We conduct %
experiments to see if a larger receptive field can help the randomly weighted network to  better capture the structured information. 
We employ the structure of `$1,1,1,1,1$' as the basic perceptual loss network and adjust the kernel size in each convolutional layer to change the receptive field. 
The results are shown in Table~\ref{tab:kernel}. 
We can see that a larger kernel size %
might 
lead to slightly better performance on average (from 70.71\% to 71.16\%).
Note that the derivation of the results increases (from 0.02\% to 0.12\%). 
%
%
%
%
%
%
%

We plot the pixel accuracy and mIoU on the validation dataset w.r.t.\ the training iterations in Figure~\ref{fig:loss}. 
Clearly the perception loss helps the training and almost during the entire training course,  we observe improved performance 
when the perception loss is used.

\begin{table}[t]
\footnotesize 
\centering
\begin{tabular}{c|c}
\hline
Kernel size & mIoU (\%)        \\
\hline
$1$           & $70.71 \pm 0.020$  \\
$3$           & $70.88 \pm 0.025$ \\
$5$           & $70.93 \pm 0.191$ \\
$7$           & $71.16 \pm 0.122$\\
\hline
\end{tabular}
\vspace{0.5em}
\caption{The perceptual loss with different kernel sizes in the perceptual loss network with five layers. With a larger receptive field, the perceptual loss %
works slightly better, with increased computation in training. 
}
\label{tab:kernel}
\vspace{-1.5em}
\end{table}

\noindent \textbf{Impact of multi-level losses.} 
A few previous works %
pay attention to aggregate multi-level losses for the perceptual loss.
We also conduct experiments to see if combining multi-level losses is helpful in our method. 
We employ the same baseline which %
achieves $69.60\%$ of mIoU in semantic segmentation on Cityscapes and uses structure of `$1,1,1,1,1$' as the perceptual loss network. 
Adding the perceptual loss at the final layer of the random network 
alone can achieve $70.88\%$ of mIoU. 
If we add five perceptual loss from the output of each convolutional block with equal loss weight, it slightly harms the performance and achieve $70.73\%$ of mIoU. 

If we adjust the weights on the losses following~\cite{wang2018high} (\textit{i.e.}, using scales of  `$\nicefrac{1}{16}$, $\nicefrac{1}{8}$, $\nicefrac{1}{4}$, 
$\nicefrac{1}{2}$, $1$' respectively), the result 
is improved slightly 
and achieves $70.93\%$ of mIoU. 

Combining multi-level of the losses 
may lead to slightly better accuracy with more hyper-parameters.  
Therefore, we only use the final layer as the generic perceptual loss in other tasks.\\

\subsubsection{%
Initialization}
In this section, we show 
the importance of initialization that we propose in Section~\ref{sec:init}. 
We  initialize the perceptual loss network with the Gaussian distributions, the uniform distributions, the Xavier-normal initializer and our developed initialization methods in Section~\ref{sec:init}, and compare their results in . 
 Table~\ref{init}.
We can see that experimental results are consistent with theoretical analysis. 

\def\+{{\color{red}{\,$\uparrow$}}}

\subsection{Dense Prediction Results}
We show that %
the generic perceptual loss can work well in different tasks
by taking the 
structured information into account during training. 
The perceptual loss network in this section refer to the %
VGG structure with $16$ convolutional layers and $5$ pooling layers, and the number of the input channel equals to the number of the task output channel. 
In semantic segmentation, the output channel equals to the number of class.
In depth estimation and instance segmentation, the output channel %
is $1$. 
The weight of the perceptual loss is set to $0.1$.

\subsubsection{Semantic Segmentation}
\label{sec:seg}
\noindent\textbf{Experiment settings.} Experiments are conducted on three benchmarks, Cityscapes~\cite{Cordts2016Cityscapes}, Pascal VOC~\cite{everingham2010pascal} and ADE20K~\cite{zhou2017scene}. 
On Cityscapes/Pascal VOC/ADE20K, the segmentation networks %
are trained by %
stochastic gradient descent (SGD) for 40K/20K/80K  epochs with $8$/$16$/$16$ training samples in the mini-batch, respectively. The learning rate is initialized as $0.01$ and is multiplied by $(1-\frac{\rm iter}{ \rm maxiter})^{0.9}$. We randomly %
crop 
the images into $769 \times 769$, $512 \times 512$, $512 \times 512$ on  these three datasets. 
Random scaling and random flipping are applied during training.

\begin{table}[t]
\small 
\centering
\begin{tabular}{l| c}
\hline
Init.\  scheme     %
& mIoU (\%) \\
\hline
${\cal N}(0,1)$                       & $ -$ \\
${\cal N}(0,0.1) $               & $45.6$ \\
${\cal N}(0,0.01)$              & $68.5$ \\
${\cal U}[-1,1] $                & $ -$  \\
${\cal U}[-0.1,0.1] $            & $51.7$ \\
${\cal U}[-0.01,0.01] $        & $69.2$ \\
Xavier-normal        & $69.8$ \\
\hline
Ours                         & $\bf{71.3}$\\
\hline
\end{tabular}
\vspace{1em}
\caption{
Results with a few different initialization  schemes. 
$\cal N$($\mu$, $\sigma$) represents  the 
Gaussian distributions with mean of $\mu$ and a stander deviation of $\sigma$. $\cal U$[$a$, $b$] represent a uniform distribution.
$ - $ means that the network fails to converge. 
}
\vspace{-1.5em}
\label{init}
\end{table}

\begin{table*}[ht]
\footnotesize
\centering
\begin{tabular}{l|p{1.5cm}<{\centering} | p{1.5cm}<{\centering} p{1.5cm}<{\centering} p{1.5cm}<{\centering}|p{1.5cm}<{\centering} p{1.5cm}<{\centering} p{1.5cm}<{\centering}}
\hline
\multicolumn{1}{c|}{\multirow{2}{*}{Method}} & \multicolumn{1}{c|}{\multirow{2}{*}{Percep.\ }} & \multicolumn{3}{c|}{Box} & \multicolumn{3}{c}{Mask} \\ \cline{3-8} 
\multicolumn{1}{c|}{}                        & \multicolumn{1}{c|}{}                        & AP     & AP$_{50}$   & AP$_{75}$   & AP      & AP$_{50}$   & AP$_{75}$   \\
 \hline
CondInst~\cite{tian2020conditional}                                      &                                              & 36.91  & 55.29  & 39.94  & 33.42   & 53.00  & 35.56  \\ %
CondInst~\cite{tian2020conditional}                                      & \checkmark                  & \textbf{37.44}  &\textbf{ 55.69}  & \textbf{40.51}  &\textbf{ 33.69}   & \textbf{53.33}  & \textbf{35.82}  \\ \hline
\end{tabular}
\vspace{0.5em}
\caption{%
Results of the generic perceptual loss for instance segmentation. Both the detection results (Box AP (\%)) and the segmentation results (Mask AP (\%)) are improved by applying the generic perceptual loss.}
\label{ins_seg}
\vspace{-2em}
\end{table*}

\noindent\textbf{Experimental results}. 
We employ three popular segmentation models with different model sizes, including a PSPNet~\cite{zhao2017pyramid} with ResNet$18$ as backbone (PSPRes18), a light-weight HRnet~\cite{wang2020deep} with $18$ layers (HRNetw18s) and the DeepLabV3+~\cite{chen2018encoder} model with ResNet50 as the backbone.
The corresponding soft targets are generated by the same architecture, but the backbone is replaced with ResNet101 or HRNet with $48$ layers. 
The baseline models are trained with the cross-entropy loss, and we further add the perceptual loss initialized with random weights. 

Table~\ref{seg_percep} %
reports the results. 
We can see that %
inclusion of 
the randomized perceptual loss can improve the performance over different datasets from $0.21\%$ to $1.6\%$, and it also works with different architectures.
If the unary term (pixel-wise loss) %
works 
sufficiently well, then  additional pair-wise or other high-order loss 
becomes less useful. 

\begin{table}[htp]
\footnotesize 
\centering
\begin{tabular}{ r |c|c| l }
\hline
Network            & Param. &Percep.               & mIoU                                                   \\ \hline
             \multicolumn{4}{c}{Cityscapes}      \\ \hline 
PSPRes18~\cite{zhao2017pyramid}             &22.9M&            & 69.6\%                                         \\
PSPRes18~\cite{zhao2017pyramid}             & 22.9M&\checkmark                 & 71.2\%   (\+1.6\%)                                          \\
HRNetw18s~\cite{wang2018high}             &3.76M&            & 73.61\%                                         \\
HRNetw18s~\cite{wang2018high}             &3.76M &\checkmark                 & 74.17\%   (\+0.56\%)\\
DeepLabV3+~\cite{chen2018encoder}          & 39.3M  &           &80.09 \%                                         \\
DeepLabV3+~\cite{chen2018encoder}          & 39.3M  & \checkmark                 & 80.70\%   (\+0.61\%)
\\\hline
             \multicolumn{4}{c}{ADE20K}      \\ \hline 

PSPRes18 ~\cite{zhao2017pyramid}             & 23.0M &                  & 33.8\%                                            \\
PSPRes18~\cite{zhao2017pyramid}             & 23.0M&\checkmark                   & 34.2\% (\+0.4\%)                                                     \\
HRNetw18s~\cite{wang2018high}             & 3.79M &                  & 31.38\%                                              \\
HRNetw18s~\cite{wang2018high}             &3.79M &\checkmark                   & 32.26\% (\+0.88\%)                              \\
DeepLabV3+~\cite{chen2018encoder}  &         39.4M  &        & 42.72\%                       \\
DeepLabV3+~\cite{chen2018encoder}  &39.4M&\checkmark                    & 42.95\%  (\+0.23\%)                      \\ \hline
             \multicolumn{4}{c}{ PascalVOC }      \\ \hline 
PSPRes18~\cite{zhao2017pyramid}             & 22.9M &                 & 49.1\%                                                      \\
PSPRes18~\cite{zhao2017pyramid}             &22.9M &\checkmark                    & 50.31\% (\+1.21\%)                                               \\
HRNetw18s~\cite{wang2018high}             & 3.76M &                  & 65.20\%                                                  \\
HRNetw18s~\cite{wang2018high}             & 3.76M&\checkmark                   & 65.41\%    (\+0.21\%)                        \\
DeepLabV3+~\cite{chen2018encoder}  &  39.3M         &        & 75.93\%                        \\
DeepLabV3+~\cite{chen2018encoder}  &39.3M&\checkmark                    & 76.79\% (\+0.86\%)                   \\\hline

\end{tabular}
\vspace{1em}
\caption{Results of semantic segmentation on three 
datasets.
\checkmark means that we employ the perceptual loss during training. 
The perceptual loss consistently improves the baseline 
across different datasets with different network structures.}
\label{seg_percep}
\vspace{-2em}
\end{table}

\subsubsection{Depth Estimation}
Depth estimation is a typical per-pixel regression problem.
We employ a plain ResNet$50$ as the backbone for depth estimation. 
The experiments are conducted on the NYUDV2 dataset~\cite{Silberman:ECCV12}.
The input images are cropped into the resolution of $385 \times 385$. 
The base learning rate is set to $0.0001$. 
We train our model using SGD with a mini-batch size of 8 for 30 epochs. 
We employ a pixel-wise weighted cross-entropy~\cite{cao2017estimating} and the vitural normal loss \cite{wei2019enforcing}.  

Predictions are evaluated by the relative error.  
The results are shown in Table~\ref{tab:depth}.
Although the VNL have already considered the geometry information 
to some extent, 
the randomized perceptual loss network can still
show 
further improvement.

\begin{table}[t]
\footnotesize 
\centering
\begin{tabular}{c|c|c|c|c}
\hline
Methods  & WCE~\cite{cao2017estimating} & VNL~\cite{wei2019enforcing} & Percep. & Rel. (\%)  \\
  \hline
a & \checkmark  &    &       & 14.5 \\
b & \checkmark   &    & \checkmark      & 14.0 \\
c & \checkmark   & \checkmark   &      & 13.6 \\
d & \checkmark   & \checkmark   & \checkmark      & \textbf{13.2}\\
\hline
\end{tabular}
\vspace{0.5em}
\caption{Depth estimation results. The generic perceptual loss complements the previous virtual normal loss~\cite{wei2019enforcing}.}
\label{tab:depth}
\vspace{-2em}
\end{table}

\subsubsection{Instance Segmentation}

For instance segmentation, we apply the generic perceptual loss based on the open source framework AdelaiDet\footnote{\url{https://git.io/AdelaiDet}}. 
We employ the state-of-the-art method CondInst~\cite{tian2020conditional} as a strong baseline. 
ResNet$50$ is %
used 
as the backbone network for CondInst, and experiments are conducted on 
the MS
COCO dataset.

Following~\cite{tian2020conditional}, models are trained with SGD on 4 V100 GPUs for 90K iterations with the initial learning rate being 0.01 and a mini-batch of 8 images. 
Other training details the same as~\cite{tian2020conditional}.
The learning rate is reduced by a factor of 10 at iteration 60K and 80K, respectively. 
Weight decay and momentum are set to  0.0001 and 0.9, respectively. 
The experiment results are %
reported 
in Table~\ref{ins_seg}.  
As bounding box detection is naturally a byproduct of CondInst, we compare results on both detection and instance segmentation tasks.
With the generic perceptual loss, the performance of both tasks are improved on all metrics.
This experiment %
again demonstrates  
that the proposed loss can benefit the 
\textit{object detection} task, when a mask branch as in CondInst is added to an one-stage object detection method.

\section{Conclusion}
In this work, we have extended  the widely used perceptual loss in image synthesis tasks to structured output learning tasks, and shows its usefulness. 
We argue that the randomly-weighted network can capture vital information among different spatial locations of the structural predictions.  
On a few image understanding tasks, including semantic segmentation, monocular depth 
estimation and instance segmentation, we demonstrate that the inclusion of this simple  
perceptual loss consistently improves accuracy.  
The proposed loss can be effortlessly  applied to many dense prediction tasks in computer vision. The only cost is the extra computation overhead during training and 
inference complexity remains the same. 
Considering its simplicity and promising performance gain, we wish to see a wide application of this generic perceptual loss in computer vision.

{\small
\bibliographystyle{ieee_fullname}
\bibliography{main}
}

\end{document}